\documentclass[]{spie}  

\def\BibTeX{{\rm B\kern-.05em{\sc i\kern-.025em b}\kern-.08em
    T\kern-.1667em\lower.7ex\hbox{E}\kern-.125emX}}
    
\usepackage{graphicx}
\usepackage{wrapfig} 
\usepackage{amssymb}
\usepackage{amsmath}
\usepackage{amsfonts}
\usepackage{booktabs}
\usepackage{url}
\usepackage{comment}
\usepackage{subcaption}
\usepackage{afterpage}
\usepackage{array}
\usepackage{float}
\everymath{\displaystyle}
\begin{document}

\title{ Leveraging synthetic imagery for collision-at-sea avoidance}
\author{Chris M. Ward, Josh Harguess, Alexander G. Corelli
\\
Space and Naval Warfare Systems Center Pacific 
\\
53560 Hull Street, San Diego, CA 92152-5001
\\
\{chris.m.ward1, alexander.corelli, joshua.harguess\}@navy.mil
}

\maketitle
\begin{abstract}
Maritime collisions involving multiple ships are considered rare, but in 2017 several United States Navy vessels were involved in fatal at-sea collisions that resulted in the death of seventeen American Servicemembers. The experimentation introduced in this paper is a direct response to these incidents. We propose a shipboard Collision-At-Sea avoidance system, based on video image processing, that will help ensure the safe stationing and navigation of maritime vessels. Our system leverages a convolutional neural network trained on synthetic maritime imagery in order to detect nearby vessels within a scene, perform heading analysis of detected vessels, and provide an alert in the presence of an inbound vessel. Additionally, we present the Navigational Hazards - Synthetic (NAVHAZ-Synthetic) dataset. This dataset, is comprised of one million annotated images of ten vessel classes observed from virtual vessel-mounted cameras, as well as a human ``Topside Lookout" perspective. NAVHAZ-Synthetic includes imagery displaying varying sea-states, lighting conditions, and optical degradations such as fog, sea-spray, and salt-accumulation. We present our results on the use of synthetic imagery in a computer vision based collision-at-sea warning system with promising performance.

\end{abstract}

\section{Introduction} \label{sec:intro}
Availability of well-annotated data is a critical component in the successful application of modern machine learning techniques to a problem domain. We seek to leverage openly available tools to generate synthetic data that can be used to train models in problem domains with sparse data. In this paper, we focus on the problem of maritime-traffic analysis as an experimentation vehicle.

Section \ref{sec:previous} provides a brief background on the work and utility of using synthetically generated imagery in machine learning applications. Section \ref{sec:navhaz} of this paper introduces the synthetic dataset we created for this study, and Section \ref{sec:experimentation} details the use of our synthetic data applied to the problem of martime collision-avoidance. We conclude the paper in section \ref{sec:conclusion}, summarizing the results of the study and proposed future work.

\section{Previous Work}\label{sec:previous}
Synthetic imagery has been used before across a wide variety of problem spaces as a way to get past the limitations of having small real-world annotated datasets.  As an example, the Virtual KITTI \cite{gaidon2016virtual} and Synthia \cite{ros2016synthia} datasets have demonstrated that for vehicular-focused imagery systems, incorporation of synthetic data outperforms using only real-world data. \cite{gaidon2016virtual,ros2016synthia,handa2015synthcam3d,hattori2015learning,marin2010learning,shafaei2016play,de2017procedural}. It has also been shown that using only synthetic data yields results on par with using a limited set of real-world training data\cite{handa2015synthcam3d,hattori2015learning,shafaei2016play}. Additionally, in G. Ros et al.\cite{ros2016synthia} it is demonstrated that the combination of synthetic data with real-world data outperforms using either set exclusively. de Souza et al.\cite{de2017procedural} further emphasizes this claim, stating that, at least for their datasets, increasing the amount of the real-world data to match that of the synthetic outperforms other models. The best results came from training on a mix of real-world and synthetic data, and then fine-tuning with examples from the target dataset \cite{ros2016synthia,shafaei2016play}. Fine-tuning using the target data helps the model account for the realities of the given domain, such as image noise, that might not be fully accounted for in the synthetic data.\cite{tremblay2018training}
\goodbreak

\section{Navigational Hazards - Synthetic (NAVHAZ-Synthetic) dataset} \label{sec:navhaz}

In this section we introduce the Navigational Hazards - Synthetic (NAVHAZ-Synthetic) dataset, and describe our methodology for creating it. This dataset is comprised of one-million labeled images of ten generic vessel classes, observed from ten shipboard points. In order to capture the high level of scene variance in maritime imagery, NAVHAZ-Synthetic contains images with diverse sea-states, weather conditions, lighting conditions, sensor noise, and optical degradations including sea-spray, and salt-accumulation. Each image contains a single target vessel and is labeled with the class and heading.

\[ \Big|\text{Classes}\Big| * \Big|\text{Vessel Headings}\Big| * \Big|\text{Sun Positions}\Big| * \Big|\text{Sky Conditions}\Big| * \Big|\text{Sea State}\Big| 
* \Big|\text{Observers}\Big| = 1x10^6\text{ images}\]

\subsection{Unity gaming engine}\label{sec:unity}
Based on work in using video games a basis for training machine learning models\cite{shafaei2016play}, we selected the Unity game engine as our modeling environment. Unity provides several advantages over proprietary modeling solutions: it is free, it is has a strong online support community, and it supports a wide variety of model formats and platforms. Running on an Intel i7 notebook, we used the Unity engine to generate the NAVHAZ-Synthetic dataset at a rate of 100,000 images/hour.

\subsection{Classes}\label{sec:vessels} 
NAVHAZ-Synthetic is comprised of ten commercial vessel classes: Liquid natural gas (LNG) Tanker, Oil Tanker, Container Ship 1 (Ballast Condition), Container Ship 1 (Full Load), Container Ship 2 (Ballast Condition), Container Ship 2 (Full Load), Cargo Ship 1, Cargo Ship 2, Cargo Ship 3, and Barge. While we did not perform any class recognition in our experimentation, each image in NAVHAZ-Synthetic is labeled with its respective vessel class.

\subsection{Light Source}\label{sec:timeofday} 
\begin{figure}[H]
\captionsetup{skip=.5cm}
\begin{center}
\captionsetup{justification=centering}
\includegraphics[width=100mm,scale=1]{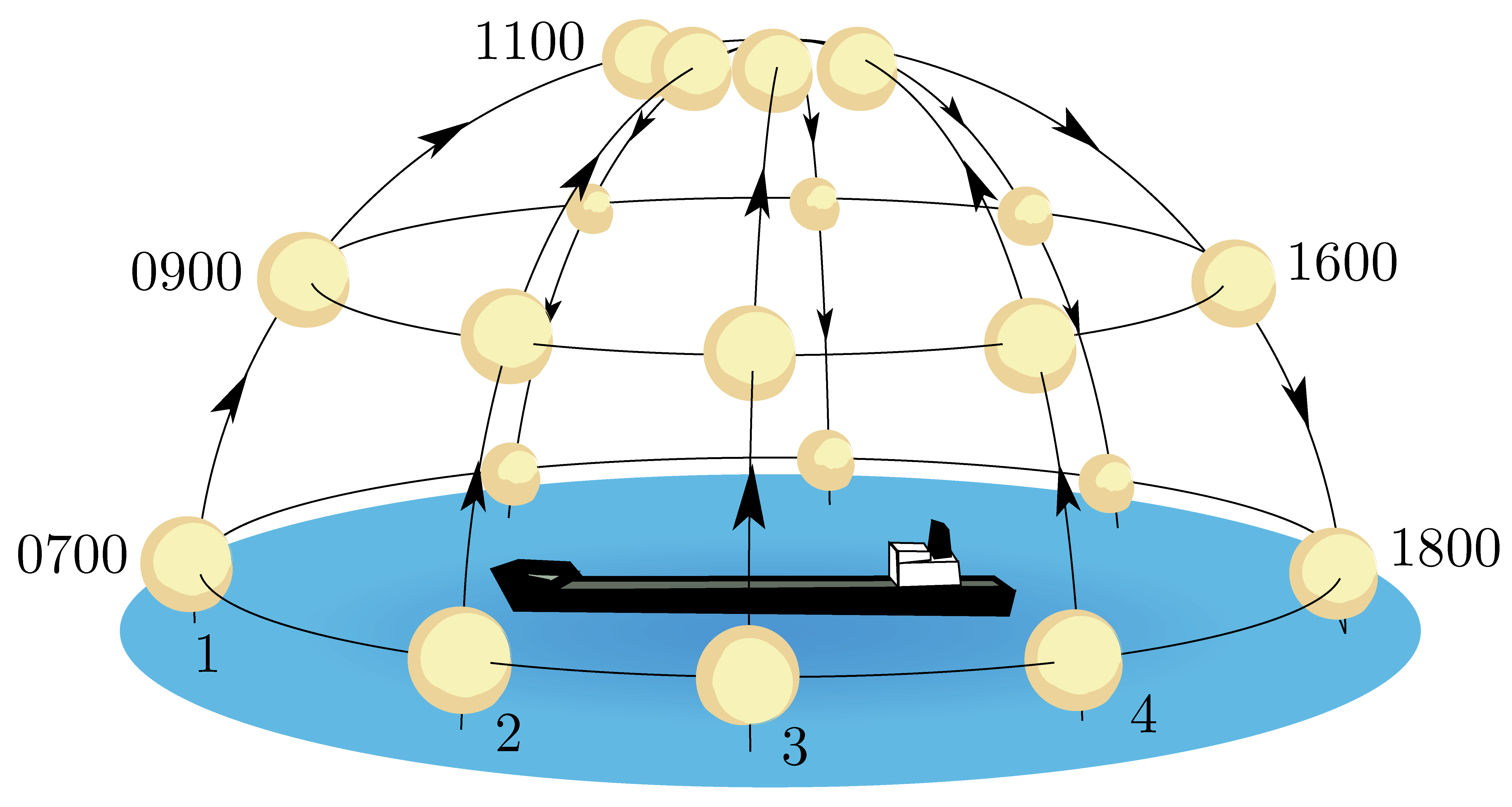}
\caption{Scene lighting positions \label{fig:sunpositions}} 
\end{center}
\end{figure}

Lighting conditions were varied over four hemispherical paths, and five positions along each path, as depicted in Figure \ref{fig:sunpositions}. We selected light-source positions that represent four daytime intervals: dawn, morning, noon, afternoon, and dusk. Additionally, we programmatically vary sky coloration to simulate light scattering in dawn and dusk conditions. \\

\hfill

\pagebreak

\subsection{Vessel Headings}
\begin{figure}[h!]
\captionsetup{skip=.5cm}
\begin{center}
\captionsetup{justification=centering}
\includegraphics[width=73mm,scale=1]{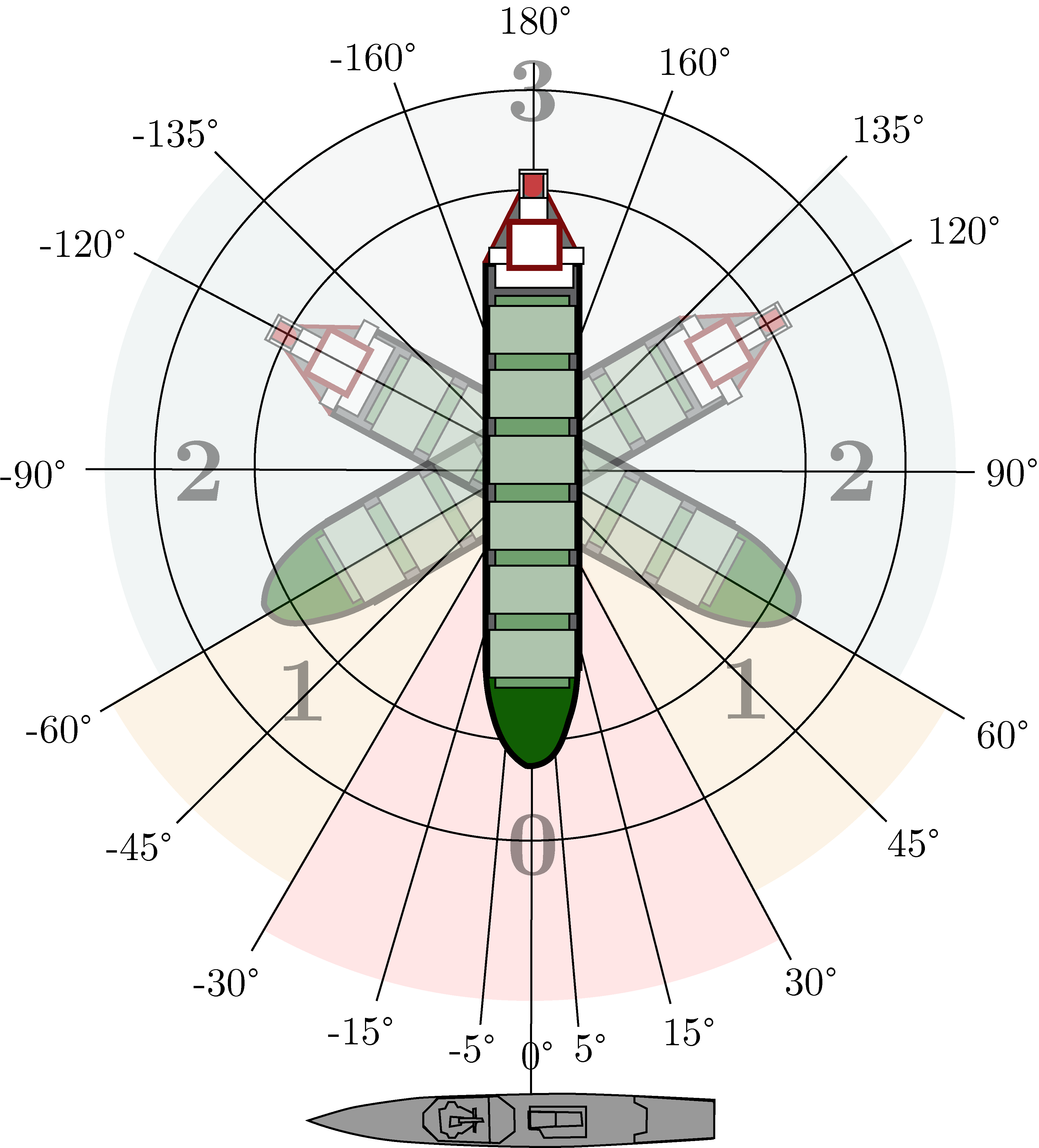}
\caption{Target vessel heading angles\label{fig:headings} }
\end{center}
\end{figure}

Each vessel class was rendered at twenty different headings as depicted in Figure \ref{fig:headings}:
\begin{center} 
$(0^{\circ}, \pm 5^{\circ}, \pm15^{\circ}, \pm30^{\circ}, \pm45^{\circ}, \pm60^{\circ}, \pm90^{\circ}, \pm120^{\circ}, \pm135^{\circ}, \pm160^{\circ}, 180^{\circ})$ 
\end{center}
Each image is labeled with the heading of the target vessel. In our experimentation we group the vessel headings into specific zones. These groupings are further explained in Section \ref{sec:experimentation}.

\subsection{ Weather conditions and sea state}\label{sec:environmentalConditions}
\begin{figure}[H]
\captionsetup{justification=centering}
\begin{subfigure}{.19\textwidth}
\includegraphics[width=\linewidth]{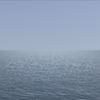}
\caption{\scriptsize Sea State: \textbf{2}\linebreak Wave height: 0.1m-0.5m\linebreak \textbf{Smooth}}
\end{subfigure}\hfill
\begin{subfigure}{.19\textwidth}
\includegraphics[width=\linewidth]{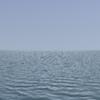}
\caption{\scriptsize Sea State: \textbf{3}\linebreak Wave height: 0.5m-1.25m\linebreak \textbf{Slight}}
\end{subfigure}\hfill
\begin{subfigure}{.19\textwidth}
\includegraphics[width=\linewidth]{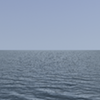}
\caption{\scriptsize Sea State: \textbf{4}\linebreak Wave height: 1.25m-2.5m\linebreak \textbf{Moderate}}
\end{subfigure}\hfill
\begin{subfigure}{.19\textwidth}
\includegraphics[width=\linewidth]{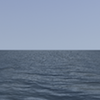}
\caption{\scriptsize Sea State: \textbf{5}\linebreak Wave height: 2.5m-4mm\linebreak \textbf{Rough}}
\end{subfigure}\hfill
\begin{subfigure}{.19\textwidth}
\includegraphics[width=\linewidth]{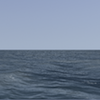}
\caption{\scriptsize Sea State: \textbf{6}\linebreak Wave height: 4m-6m\linebreak \textbf{Very rough}}
\end{subfigure}
\caption{Sea States\label{fig:seastates}}
\end{figure}

Environmental conditions at sea are highly variant. Leveraging the power of synthetic imagery, we attempted to capture this variance by training on a different combination of weather and sea conditions. NAVHAZ-Synthetic features five weather conditions: clear skies, dynamic clouds, dense dynamic clouds, overcast sky, and dense fog layer.
Additionally, we selected five sea states (shown in Figure \ref{fig:seastates} ) that correlate to conditions described by the World Meteorological Organization Sea State codes\cite{bales1983designing}. Combined, these weather and sea state variations yield twenty-five environmental conditions across the training data.

\newpage
\subsubsection{Observers} \label{sec:observers}
\begin{figure}[h!] 
\begin{center}
\captionsetup{justification=centering}
\includegraphics[width=80mm,scale=.8]{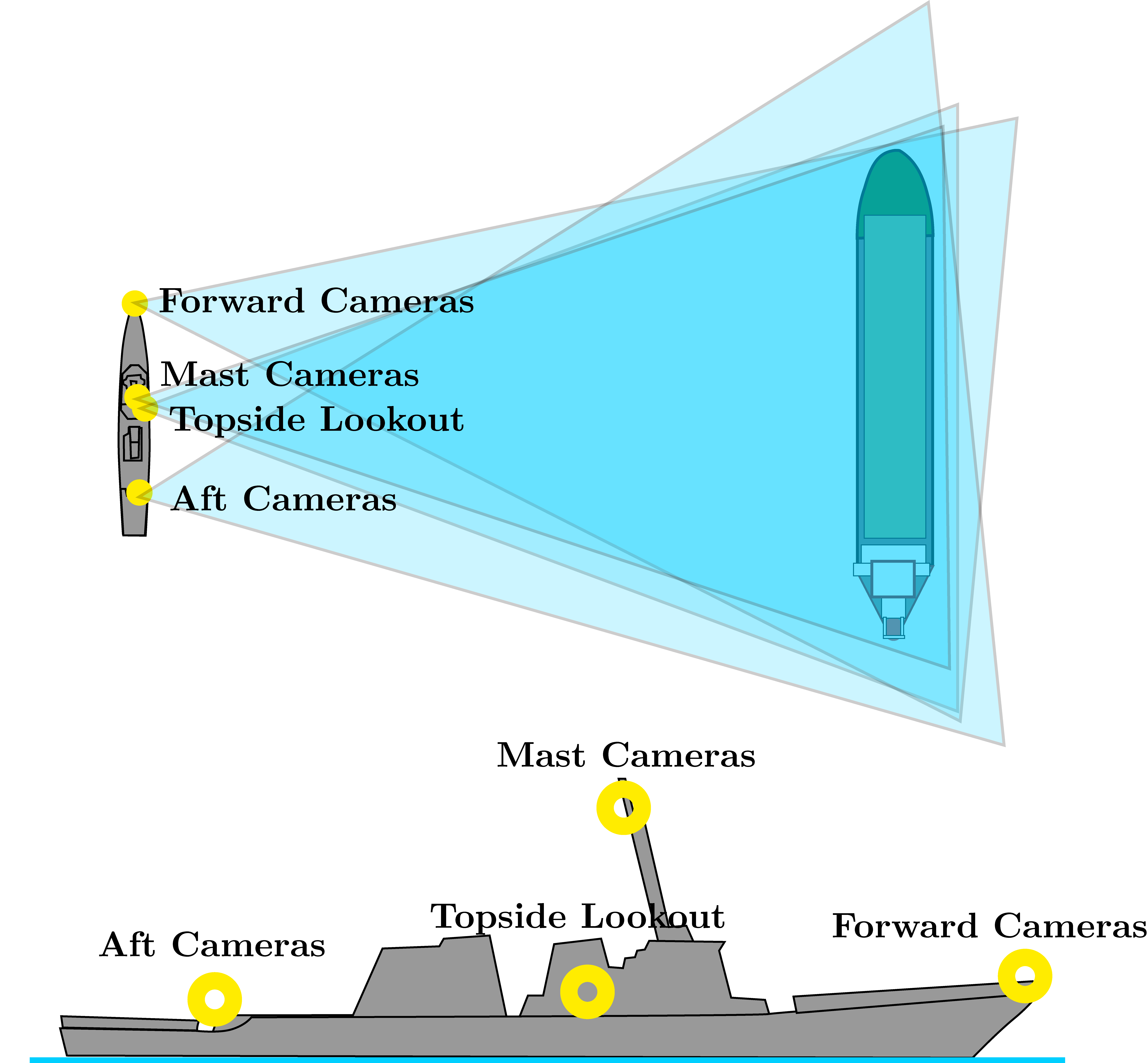}
\caption{Observer Positions\label{fig:observerpositions}}
\end{center}
\end{figure}

\begin{figure}[ht]
\captionsetup{justification=centering}

\begin{subfigure}{.28\textwidth}
\includegraphics[width=\linewidth]{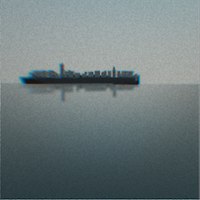}
\caption{\scriptsize Forward camera 1}
\end{subfigure}\hfill
\begin{subfigure}{.28\textwidth}
\includegraphics[width=\linewidth]{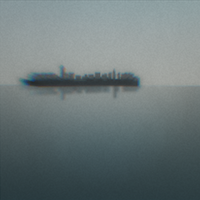}
\caption{\scriptsize Forward camera 2}
\end{subfigure}\hfill
\begin{subfigure}{.28\textwidth}
\includegraphics[width=\linewidth]{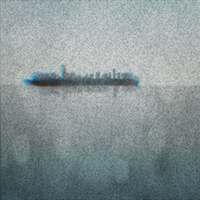}
\caption{\scriptsize Forward camera 3}
\end{subfigure}\par

\begin{subfigure}{.28\textwidth}
\includegraphics[width=\linewidth]{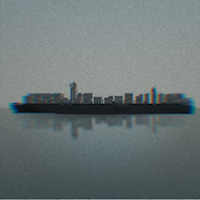}
\caption{\scriptsize Mast camera 1}
\end{subfigure}\hfill
\begin{subfigure}{.28\textwidth}
\includegraphics[width=\linewidth]{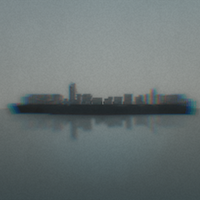}
\caption{\scriptsize Mast camera 2}
\end{subfigure}\hfill
\begin{subfigure}{.28\textwidth}
\includegraphics[width=\linewidth]{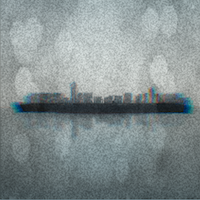}
\caption{\scriptsize Mast camera 3}
\end{subfigure}\par

\begin{subfigure}{.28\textwidth}
\includegraphics[width=\linewidth]{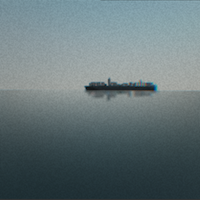}
\caption{\scriptsize Aft camera 1}
\end{subfigure}\hfill
\begin{subfigure}{.28\textwidth}
\includegraphics[width=\linewidth]{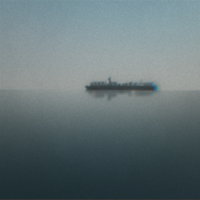}
\caption{\scriptsize Aft camera 2}
\end{subfigure}\hfill
\begin{subfigure}{.28\textwidth}
\includegraphics[width=\linewidth]{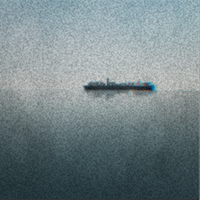}
\caption{\scriptsize Aft camera 3}
\end{subfigure}\par


\hfill\linebreak
\caption{Scene Observers}
\label{fig:cameraobservers}
\end{figure}

In order to generate sufficient observational diversity, we virtualized ten observers in the modeling environment. It is common to have a human topside lookout in maritime evolutions, thus we included such an observer in our Unity environment. A topside lookout was modeled with a field-of-view that resembles the perspective a human observer with binoculars. The remaining nine observers represent a ship-mounted camera system augmented with degradations and effects inherent in real-world maritime conditions. The camera observers were modeled using common parameters from commercially available maritime cameras and placed at various points on ownship as depicted in Figure \ref{fig:observerpositions}. We visualize the synthetic camera feeds in Figure \ref{fig:cameraobservers} and show the effects and degradations associated with each observer in Table 1.
\hfill

\begin{table}[H]\label{tab:observerParams}
\begin{center}
\begin{tabular}{|>{\centering}m{3cm}|>{\centering}m{1cm}| >{\centering}m{1.25cm}|>{\centering}m{2cm}|>{\centering}m{1.25cm} |>{\centering}m{1.25cm}|>{\centering}m{1.25cm} |c |}
\hline
\textbf{Observer Name} & \textbf{Field of View} & \textbf{Focal Length} & \textbf{Chromatic Aberration} & \textbf{Lens Flare} & \textbf{Lens Bloom} & \textbf{Lens Grit} & \textbf{Sensor Grain}\\
\hline

Topside Lookout  & 31.0 & 54.7 & - & - & - & - & -\\
\hline
Forward Camera 1  & 33.2 & 58.7 & \checkmark & -  & - & - & \checkmark\\
Forward Camera 2  & 33.2 & 58.7 & \checkmark & - & \checkmark & \checkmark & \checkmark\\
Forward Camera 3  & 33.2 & 58.7 & \checkmark & \checkmark & \checkmark & \checkmark\checkmark & \checkmark\checkmark\\
\hline
Mast Camera 1  & 22.0 & 58.7 & \checkmark & - & - & - & \checkmark\\
Mast Camera 2  & 22.0 & 58.7 & \checkmark & \checkmark & \checkmark & \checkmark & \checkmark\\
Mast Camera 3  & 22.0 & 58.7 & \checkmark & \checkmark & \checkmark & \checkmark\checkmark & \checkmark\checkmark\\
\hline
Aft Camera 1  & 53.1 & 35 & \checkmark & - & - & - & \checkmark\\
Aft Camera 2  & 53.1 & 35 & \checkmark & \checkmark & \checkmark & \checkmark & \checkmark\\
Aft Camera 3  & 53.1 & 35 & \checkmark & \checkmark & \checkmark & \checkmark\checkmark & \checkmark\checkmark\\
\hline
\end{tabular}
\hfill
\caption{Observer Parameters}
\end{center}
\end{table}

\newpage

\section{Experimentation} \label{sec:experimentation}
This section details our experimentation. We create a very large synthetic dataset (1-million images) and use it to train a deep neural network for a classification task: determining the heading of a commercial vessel. We evaluate our ability to train a large capacity model (VGG-19)\cite{simonyan2014very} as well as the performance of the trained model on a small dataset of real imagery.

\subsection{Datasets}\label{sec:data}
This section gives an overview of the image datasets used in our experimentation.

\subsubsection{Training and validation data}\label{sec:trainingValidationData}
In addition to using ImageNet weights \cite{simonyan2014very, deng2009imagenet} for transfer learning, our training data consists of an 800,000 image class-balanced subset of NAVHAZ-Synthetic. We also curated a small subset (250 images) of commercial ships from the ImageNet dataset to use for domain adaptation. We used a 20\% validation holdout in both training sets. We did not perform any training data augmentation in this experiment.

\subsubsection{Test data}\label{sec:testData}
Our test data consists of a second small ImageNet subset (200 images) of commercial ships, balanced over four classes. Because it is difficult to infer the exact heading of the vessels in the ImageNet data, we apply weak labels, (general heading of the target vessel) to the test images. Our labeling schema is further described in \ref{sec:experimentDesign}.

\subsection{Architecture specification}\label{sec:vgg19}
For this experiment, we used the VGG-19 convolutional neural network described in Simonyan and Zisserman\cite{simonyan2014very}. The input layer was modified to accommodate the NAVHAZ-Synthetic dimensions of 384x384 pixels. Following the VGG-19 stack of convolutional layers, the top of our model consists of three fully-connected layers: two with 4096 channels, followed by a 4-channel fully connected layer that corresponds to our four output classes. The model is finalized with a softmax output layer.

\subsection{Experiment design}\label{sec:experimentDesign}
The problem of detection was bypassed in this experiment, however, maritime detection is a non-trivial problem. To alleviate detection-performance requirements for integration of our work, we train on images without tightly cropped bounding boxes, and targets generally centered in the frame. 

Although NAVHAZ-Synthetic provides exact headings of target vessels, we weakened the labels to facilitate shaping this experiment as a classification problem, versus a more challenging regression problem. Weakening the labels also alleviates the challenge of labeling the ImageNet data with precise vessel headings. The data was grouped into four classes: Zone 0 (imminent collision), Zone 1 (collision danger), Zone 2 (safe maneuvering), and Zone 3 (no danger). These zones are visualized in Figure \ref{fig:headings}.

We trained VGG-19 in three discrete steps. First, we initialized all but the fully-connected layers of the model with ImageNet weights. We initialized the fully-connected layers with small random values chosen from a zero-mean Gaussian.\cite{hinton2012practical} To establish a basis for performance comparison, we then attempted to fine-tune the model on our small ImageNet training subset of commercial ships. 

In the second part of our experiment, we re-ininitialized the model with ImageNet weights and fine-tuned the network on our NAVHAZ-Synthetic training set. We used early stopping to achieve maximum validation accuracy without overfitting. We then tested our model on the ImageNet test set.

In the third part of our experimentation we performed domain adaptation by fine-tuning the model again on the ImageNet training set. We then retested our model on the ImageNet test set.

\subsection{Experimental results} \label{sec:results}
When training on just 250 real images we expected to see VGG-19 overfit, yielding a poorly-generalized model. However, dropout in the fully-connected layers is, by design, set to a rate of 50\% to prevent such overfitting. We were unable to achieve convergence with such a small training set. As such, this base-performance comparison is omitted from our analysis.

\begin{figure}[H]
\captionsetup{justification=centering}
\begin{subfigure}{.45\textwidth}
\includegraphics[width=\linewidth]{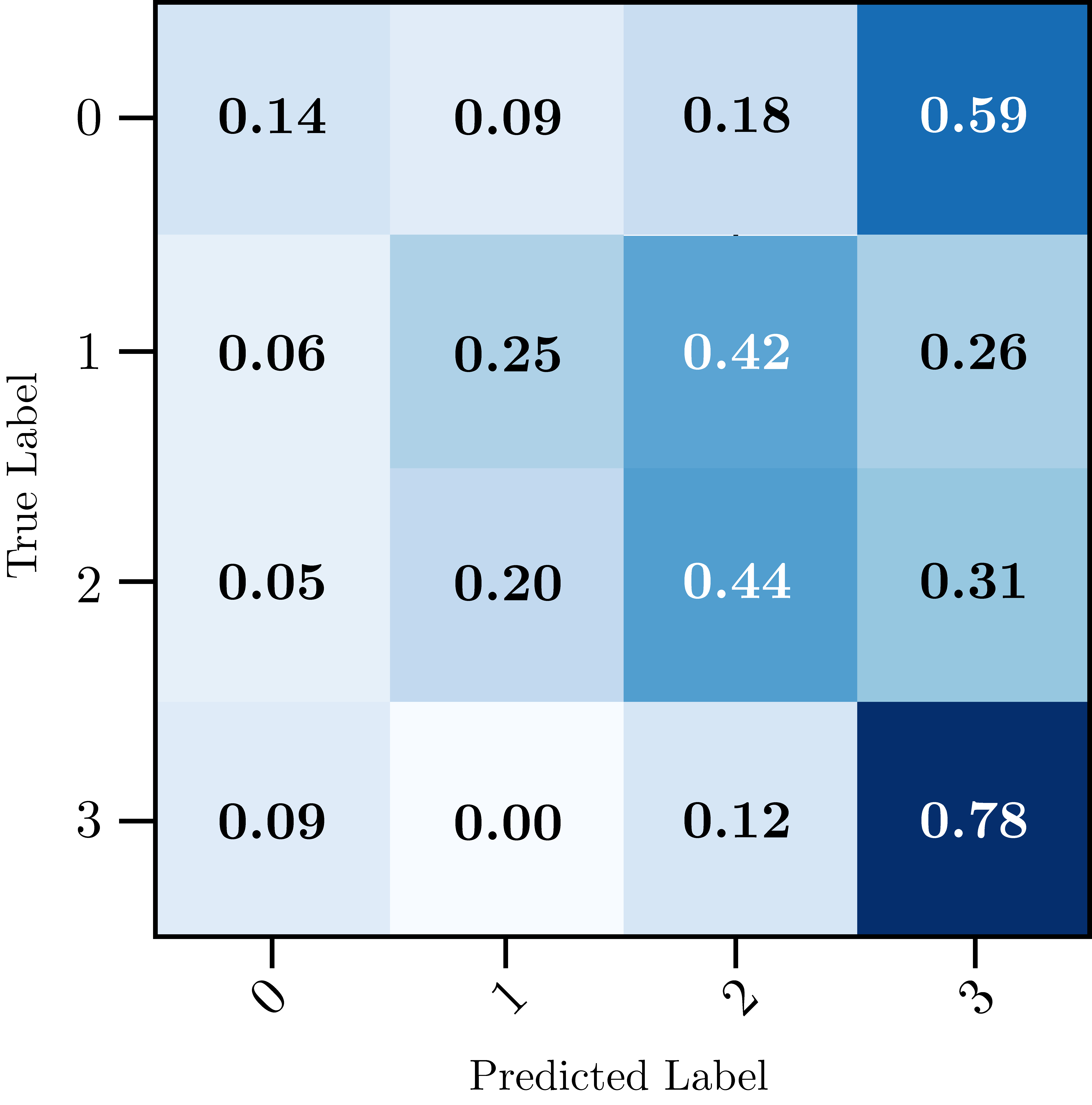}
\caption{Synthetic-only training applied to real data\label{fig:sytheticonlyresults}}
\end{subfigure}\hfill
\begin{subfigure}{.45\textwidth}
\includegraphics[width=\linewidth]{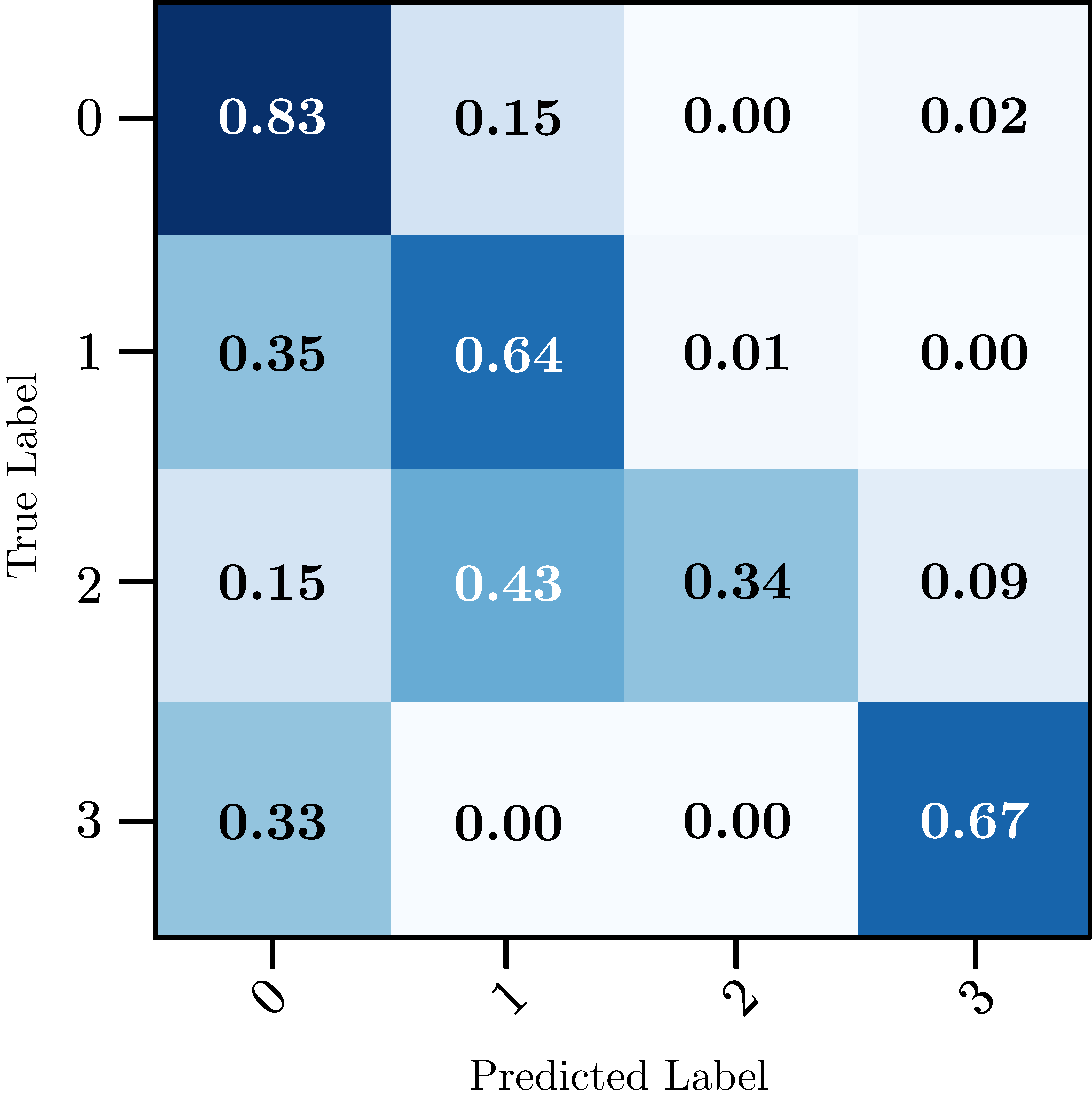}
\caption{Domain adaptation results \label{fig:adaptationresults}}
\end{subfigure}
\caption{Normalized confusion matrices\label{fig:confusion}}
\end{figure}

When trained on fully-synthetic data we achieved a high validation accuracy (3\% error), but poor test accuracy (70\% error). We can see from the resulting confusion matrix (Figure \ref{fig:sytheticonlyresults}) that our model performs only slightly better than a random guess when presented with real test data.

After performing domain adaptation, we see a significant increase in accuracy on the test set, reducing the classification error from 70\% to only 36.5\% after a single epoch of fine-tuning on real training data.
The confusion matrix for the domain adaptation results (Figure \ref{fig:adaptationresults}) suggests that performance may improve even more when accounting for inter-class noise, noting that vessels classified with a Zone 0 heading are visually similar to those with a Zone 1 heading.\\
\goodbreak

\section{Conclusion and Future Work} \label{sec:conclusion}
We presented NAVHAZ-Synthetic: a very large dataset of ten commercial vessel classes under varying lighting and environmental conditions. By applying this  dataset, we demonstrated that synthetic data can be used to a train complex neural network architecture to convergence. We also show that domain adaptation can be accomplished with use of real imagery, yielding a well generalized model with very little target data. This is particularly significant for maritime and other Department of Defense (DoD) problems where data is often sparse and weakly labeled. 

In leveraging synthetic imagery and constraining our problem to a simplified computer vision task, such as determining the general heading of an observed ship, we have demonstrated that current work in computer vision and machine learning can be adapted and applied to maritime problems like collision-avoidance. Alleviating  the need for fielding a complex and expensive shipboard system, our proposed method is capable of running on a commercially available computing platform using video feeds from existing camera systems. 

Statistical analysis of synthetic imagery versus real target data could help us understand how to further improve our results, although work in this domain seems to indicate that improving the ``realism" of synthetic data yields a diminishing return.\cite{mayer2018makes}. Moreover, we have only implemented a very basic approach to domain adaptation, and will explore more sophisticated methods in order to understand and maximize the efficacy of synthetic imagery in training machine learning models.

\hfill
\hfill
Please contact the authors if you are interested in acquiring the NAVHAZ-Synthetic dataset, or our VGG-19 model weights. 

\goodbreak

\clearpage
\bibliography{synth-SPIEDCS18}   
\bibliographystyle{spiebib}   

\end{document}